# LLaVaOLMoBitnet1B: Ternary LLM goes Multimodal!

Jainaveen Sundaram, Ravi Iyer
{jainaveen.sundaram, ravishankar.iyer}@intel.com

## I. Introduction

Multimodal Large Language Models (MM-LLMs) have seen significant advancements in the last year, demonstrating impressive performance across tasks. While closed source models such as GPT-4o/Claude/Gemini are topping leaderboards [1], LLaVa [2] and its variants remain some of the best open-source MM-LLMs with a strong adoption by the developer community.

To truly democratize AI, apart from strong capabilities, models must run efficiently on small compute footprints accessible by most. Small Language Models (SLMs) address this gap, where number of parameters are scaled down (typically <3B) while keeping architecture choices and pre-training token exposure close to their larger counterparts. As a result, models such as Phi [3], Gemma-2b [4] amd Olmo [5] exhibit strong performance across benchmarks with a smaller memory footprint and lower compute latency.

Weight Quantization offers an additional knob to further shrink model sizes while balancing performance. Microsoft's BitNet1.58b LLM [6] pushes weights tensors down to an impressive ternary set of {-1,0,1} with minimal performance hit. NousResearch implemented a proof-of-concept open-source implementation of [6] called OLMo-Bitnet-1B [14] - the first Ternary 1 billion paramter LLM.

We build on the work done by NousResearch to extend the capabilities of ternary models beyond text. Our contributions are:

1. Built the first Ternary Multimodal LLM capable of accepting Image(s)+Text inputs to produce coherent textual responses.
2. Open-Sourced the model along with weights and training scripts for future research into Ternary models
3. Highlighting challenges associated and opportunities present to make Ternary models mainstream.

## II. Related Works

Flamingo [7] - widely regarded as the "GPT moment for Multimodal Models" was the harbinger to rapid advances in MM-LLMs. One of the seminal successors was LLaVa, which introduced an open-source framework to train MM-LLMs. LLaVa introduced a novel method by which text only GPT was used to expand to multimodal datasets, resulting in rich instruction-tuning style multimodal data.

The model architecture was kept simple, with a pre-trained image encoder connected to a pre-trained LLM via a trainable multilayer perceptron (MLP). Image encoder from CLIP [8] was chosen to project image patches to a higher-dimensional space, which the MLP re-projects to the LLM (originally Vicuna [9]) embedding space. The LLM uses these image features along with input text embeddings to generate a response.

The versatility of LLaVa framework inspired numerous variants, with different Image encoders, LLMs, fine-tuning data mixtures and applications to domains. Small MM-LLMs were also developed using the LLaVa framework, including TinyLLaVa [10] and LLaVa-Gemma [11].

As models were getting smaller, the BitNetb1.58 work [6] proposed a novel method to quantize weights to extremely low precisions (ternary). Their work showed Post-Training Quantization introduces significant accuracy degradation at extremely low precisions, and instead pointed out that pre-training with low precision weights is a viable method to create strong ternary models. The authors presented a family of models from 1.3B to 70B parameters, providing a latency benefit of upto 4X with minimal hit to accuracy across benchmarks.

While BitNetb1.58 has not yet been open-sourced, an AI company called NousResearch independently verified their method by open-sourcing a ternary version of OLMo (called OLMoBitNet1B [14]) - where all the linear layers are replaced by a ternary BitLinear layer. Trained on 60B tokens of the Dolma dataset [13], it is still undertrained when compared to its peers. For reference, Microsoft's BitNetb1.58 was trained on 100B tokens while the full precision OLMo was trained on 2T tokens.

## III. Model details

### III.A Model Architecture

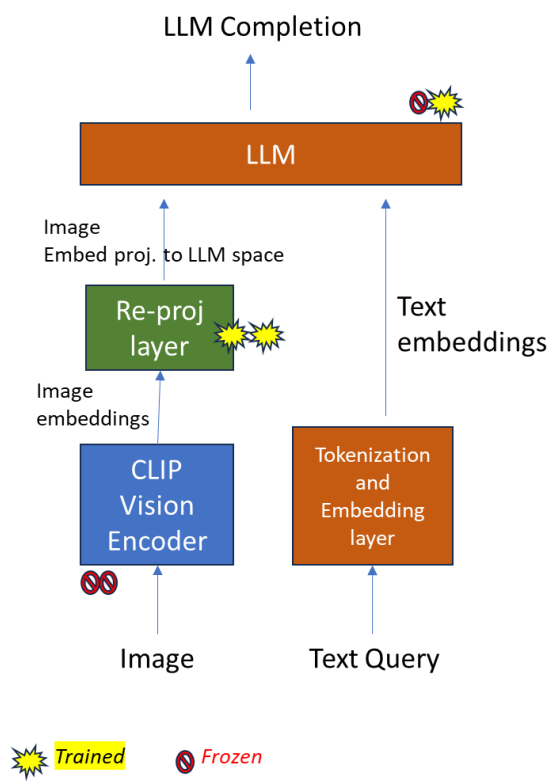

*Fig.1 Overall Model architecture and what stages are frozen/trained during each phase of training*

This section describes how we go about building LLaVaOLMoBitNet1B - a Multimodal LLM based out of Ternary OLMoBitnet1B model using the LLaVa recipie. Though the Vision Encoder and Reprojection layer maintain full precision weights, we take the liberty of calling it a Ternary model as the dominant chunk of the model weights (LLM accounting for >90% parameters) are in the ternary domain. As shown in Fig. 1, LLaVaOLMoBitNet1B is made up of three parts: (1) A CLIP ViT-L/14 type vision encoder (2) An MLP connector followed by a (3) Ternary LLM. Input image is split into non-overlapping patches of 14x14 and passed to the vision encoder. The vision encoder is made of 24 transformer layers with a hidden dimension of 1024, totaling 100M paramters.

Hence an image of size (w,h) results in: N=(w * h)/(14 * 14) patches, which are flattened and passed through the encoder. For each image, output of the encoder is (N, d), where d: 1024 (hidden dimension of the encoder).

The encoder output is then passed to an MLP, which re-projects image features to the LLM embedding space. The MLP here is made of two linear layers

(hidden dimension of 2048), sandwiching a GELU activation. Output of the MLP is of dimension (N, d'') where d': 2048, which is also the embedding dimension of the LLM.

The LLM used here is the ternary OLMoBitNet1B developed by NousResearch. The core of the LLM is made of 16 transformer decoder layers, with all the linear layers replaced by BitLinear158 layers introduced in [12]. With a total 1.1 billion parameters, OLMoBitNet1B was trained as a PoC on the first 60B tokens of the Dolma dataset [13].

Input text query is first tokenized and passed through the embedding layer of the LLM to result in a matrix of (m, d'), where m is the number of generated tokens. This matrix concatenated with the image projected tensor, resulting in an (m+n, d') tensor passed to the LLM. Conditioned on the input context, the LLM autoregressively produces response tokens based on output sampling conditions.

### III.B Training details

We follow the same training approach outlined in the LLaVa1.5 paper, consisting of two phases: (1) A Pre-training phase for feature alignment followed by an (2) End-to-end instruction fine-tuning. Details of the dataset through the two stages are shown in Fig. 2.

DeepSpeed library was used for multi-GPU training for both training phases. The pre-training phase involves 1 epoch on a filtered subset of 595K Conceptual Captions [2], with only the projection layer weights updated. Trained on a GPU cluster, we set batch size to be 32 (per device), with gradients accumulated every 4 steps. We had to use a smaller batch size than LLaVa prescription, as OLMoBitNet1B model did not support gradient checkpointing. Learning rate was set to 1e-3, with cosine decay and a warmup ratio of 0.03.

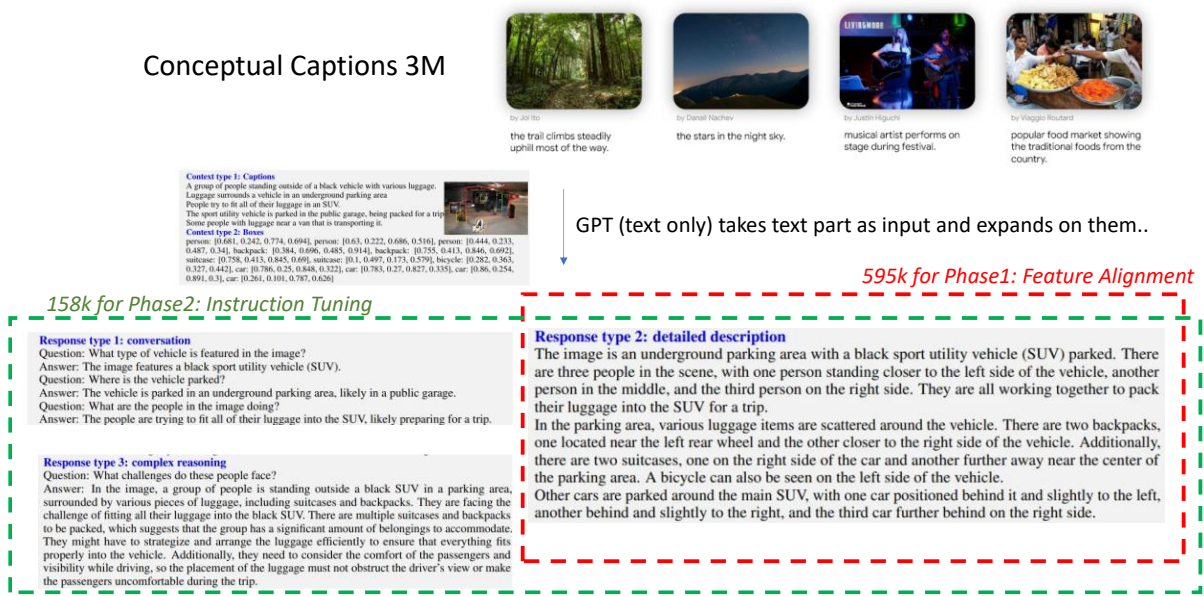

Fig2. Represents how LLaVa expands on CC dataset using GPT to create a detailed multimodal instruction dataset

For instruction fine-tuning, we use 1 epoch of the LLaVa-Instruct-150K dataset, with both projection layer and LLM weights updated. We adopt full LLM finetuning instead of any low-rank approaches. Due to larger gradient computation requirement, we drop batch size to 8, accumulating gradients every 2 steps. Learning rate is dropped to 2e-5, with similar decay and warm-up ratios as before. Adam was the chosen optimizer, with momentum parameters set to 0.9 and 0.98.

## IV. Results

We evaluate the final LLaVaOLMoBitNet1B model qualitatively and quantitatively for image + text inference performance. Fig. 3 shows qualitative results, with questions posed on input images. We see the model produce coherent and mostly correct responses with some inaccuracies. For

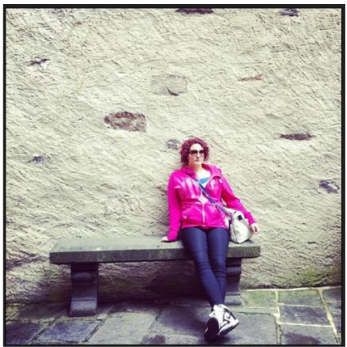

(a) Describe what is in the image:

*The image features a woman sitting on a stone bench, wearing sunglasses and a pink jacket. She is wearing sunglasses and a pink jacket, and she is positioned near a wall. The bench is located in a courtyard, and the woman is sitting on it. The setting appears to be a stone wall, possibly a part of a building or a public space.*

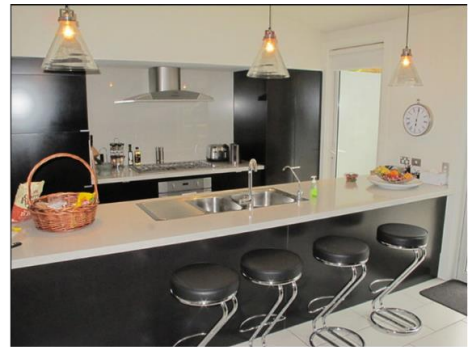

(b) Does the image contain any chairs or stools? If so, describe them and tell me how many

*Yes, there are two stools in the image, and they are black in color.*

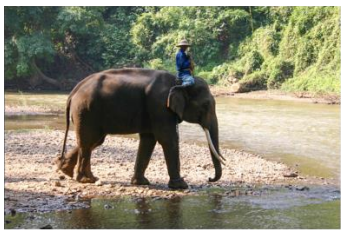

(c) Give a short description of the image.

*The image features a man riding an elephant in a river, surrounded by trees. The man is wearing a blue shirt and a hat, and the elephant is walking through the water. The man is positioned near the center of the image, while the elephant is positioned towards the right side. The man is wearing a blue shirt and a hat, and the elephant is walking through the water. The scene captures the essence of the natural environment and the bond between the man and the elephant.*

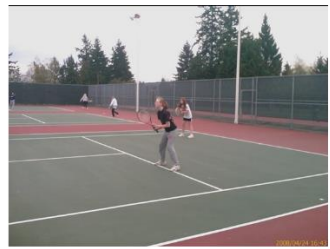

(d) Be concise. What are the major tournaments of the sport shown in the image?

*The image shows four major tournaments of the sport shown in the image: the Australian Open, the French Open, and the US Open. The players are actively engaged in a match, with one player holding a tennis racket and preparing to hit the ball. The players are actively engaged in the game, with one player holding a tennis racket and preparing to hit the ball. The presence of multiple players and the competitive atmosphere suggest that this is a competitive event.*

*Fig3. Qualitative responses produced by the model for some example text + image query*

example, in Fig. 3b the model picks 'stools' correctly, identifies its color but gets the count wrong. Similarly, in Fig. 3c the model gets the decription correct but makes errors in relative positions of entities. On the flip side, In Fig. 3d, the model misses Wimbledon altogether, and provides a verbose explanation of Tennis.

Before examining quantitative results of the MM-LLM, Table 1 compares the accuracy performance of the base LLM used (OLMoBitNet1B) against its peers against common benchmarks:

| Model | Open/Close source | Params | Pretrained tokens | Arc easy | Open bookqa | Piqa | Sciq |
|---|---|---|---|---|---|---|---|
| Gemma2b | Closed | 2B | 6T | 80.1 | | 77.8 | |
| Olmo 1b | Open Source | 1B | 2T | 58.07 | 46.4 | 73.7 | 88.1 |
| BitNetb1.58 | Closed | 1.3B | 100B | 54.9 | 19.6 | 68.8 | |
| OLMoBitNet1B | Open Source | 1B | 60B | 49.93 | 30.4 | 67.25 | 74.3 |

*Table1: Comparison of ternary OLMoBitNet1B against peers of the same size*

As seen, OlmoBitnet1B is pre-trained on the least number of tokens, and unsurprisingly exhibits lowest scores on benchmarks.

When comparing accuracy of LLaVaOLMoBitNet1B across common Multimodal benchmarks, we see a similar trend as seen in Table 1:

| Model | Params | Training method | POPE | VQAv2 | Text VQA |
|---|---|---|---|---|---|
| TinyLLaVa 3B | 3B | Llava recipie | 86.8 | 77.6 | 51.4 |
| MM1 3B | 3B | MM1 recipie | 87.4 | 82 | |
| LlaVaGemma2B | 2B | Llava recipie | 85 | 70.7 | |
| LLaVaOLMoBitNet1B | 1B | Llava recipie | 66.92 | 68.41 | 26.3 |

*Table2: Comparison of the multimodal ternary LLM LLaVaOLMoBitNet1B against its larger peers*

Being the first of its ternary multimodal LLM, LLaVaOLMoBitNet1B remains one of the smallest models with least pre-training token exposure compared to its full precision peers. While the current model is not the strongest, it provides a good baseline against which we intend to develop more capable ternary multi-models.

## V. Future Work

In the Model development world, the stronger models tend to be Open Weight (where the dataset/training methods are not publically available) or closed source. However, the current ternarization approach mandates training models from scratch – a luxury available to only a few organizations with sufficient compute budgets. We believe the highest impact research problem would be to find effective ways to Post-Training Quantize/Quantization-Aware Finetune open weight pre-trained models to ternary domain.

In addition, ternary models also come with the same set of challenges posed by regular LLMs - around biases in response, model uncertainty and hallucinations to name a few. On the Hardware side there, as [6] rightly point out, there is a gap in efficiently mapping ternary operations to derive performance benefits near theoretical limits. We seek to address a few of the challenges and share results of our research in our future work.

## VI. Acknowledgements

We thank the authors of LLaVa framework, BitNetb1.58 and NousResearch for sharing their awesome work for us to build on. We also thank members of ESL and SCHPL teams within Intel Labs for their valuable inputs.

## VII. References

1. https://lmsys.org/blog/2024-06-27-multimodal/
2. Liu, Haotian, et al. "Improved baselines with visual instruction tuning." *Proceedings of the IEEE/CVF Conference on Computer Vision and Pattern Recognition*. 2024.
3. Phi2: https://www.microsoft.com/en-us/research/blog/phi-2-the-surprising-power-of-small-language-models/
4. Team, Gemma, et al. "Gemma 2: Improving open language models at a practical size." *arXiv preprint arXiv:2408.00118* (2024).
5. Groeneveld, Dirk, et al. "Olmo: Accelerating the science of language models." *arXiv preprint arXiv:2402.00838* (2024).
6. Ma, Shuming, et al. "The era of 1-bit llms: All large language models are in 1.58 bits." *arXiv preprint arXiv:2402.17764* (2024).
7. Alayrac, Jean-Baptiste, et al. "Flamingo: a visual language model for few-shot learning." *Advances in neural information processing systems* 35 (2022): 23716-23736.
8. Radford, Alec, et al. "Learning transferable visual models from natural language supervision." *International conference on machine learning*. PMLR, 2021.
9. Zheng, Lianmin, et al. "Judging llm-as-a-judge with mt-bench and chatbot arena." *Advances in Neural Information Processing Systems* 36 (2024).
10. Zhou, Baichuan, et al. "Tinyllava: A framework of small-scale large multimodal models." *arXiv preprint arXiv:2402.14289* (2024).
11. Hinck, Musashi, et al. "LLaVA-Gemma: Accelerating Multimodal Foundation Models with a Compact Language Model." *arXiv preprint arXiv:2404.01331* (2024).
12. Wang, Hongyu, et al. "Bitnet: Scaling 1-bit transformers for large language models." *arXiv preprint arXiv:2310.11453* (2023).
13. Soldaini, Luca, et al. "Dolma: An open corpus of three trillion tokens for language model pretraining research." *arXiv preprint arXiv:2402.00159* (2024).
14. OlmoBitnet1B: https://huggingface.co/NousResearch/OLMo-Bitnet-1B